\documentclass{article}

     \usepackage{algorithm,algcompatible,amsmath}

 \usepackage[final,nonatbib]{neurips_2019}

\usepackage[utf8]{inputenc} 
\usepackage[T1]{fontenc}    
\usepackage{hyperref}       
\usepackage{url}            
\usepackage{booktabs}       
\usepackage{amsfonts}       
\usepackage{nicefrac}       
\usepackage{microtype}      
\usepackage[pdftex]{graphicx} 
\usepackage{caption}
\usepackage{subcaption}
\usepackage[inline]{enumitem}
\usepackage[backend=biber,style=ieee,sorting=nty]{biblatex}
\usepackage{authblk}
\begin{document}

\title{Global Saliency: Aggregating Saliency Maps \\
        to Assess Dataset Artefact Bias}

%

\author[1,5]{\textbf{Jacob Pfau}}
\author[1,5]{\textbf{Albert T. Young}}
\author[1,2,5]{\textbf{Maria L. Wei}}
\author[3,4,5]{\textbf{Michael J. Keiser}}

\affil[1]{\footnotesize Department of Dermatology, UCSF and Dermatology Service, SFVAMC}
\affil[2]{\footnotesize Helen Diller Family Comprehensive Cancer Center, UCSF}
\affil[3]{\footnotesize Institute for Neurodegenerative Diseases, and Bakar Computational Health Sciences Institute, UCSF}
\affil[4]{\footnotesize Department of Pharmaceutical Chemistry, Department of Bioengineering and Therapeutic Sciences, UCSF}
\affil[5]{\footnotesize \{pfau, keiser\}@keiserlab.org, \{albert.young, maria.wei\}@ucsf.edu}

\maketitle

\begin{abstract}
In high-stakes applications of machine learning models, interpretability methods provide guarantees that models are right for the right reasons. In medical imaging, saliency maps have become the standard tool for determining whether a neural model has learned relevant robust features, rather than artefactual noise. However, saliency maps are limited to local model explanation because they interpret predictions on an image-by-image basis. We propose aggregating saliency globally, using semantic segmentation masks, to provide quantitative measures of model bias across a dataset. To evaluate global saliency methods, we propose two metrics for quantifying the validity of saliency explanations. We apply the global saliency method to skin lesion diagnosis to determine the effect of artefacts, such as ink, on model bias.
\end{abstract}

\section{Introduction}

Across medical imaging tasks, convolutional neural networks (CNN) have demonstrated human-level diagnostic accuracy \cite{meta,health}. However, these models are often not robust, learning non-generalizable patterns and confounding visual artefacts with diseased tissue \cite{xray,tomog}. Such models may perform well on a validation set yet fail catastrophically when applied in new contexts. The standard method to explain model predictions is to visually, individually inspect saliency maps for each image to determine whether the most salient pixels are the most relevant for diagnosis. This practice prevents some unexpected failures but remains subjective based on the reviewer's judgement. Hence, we expand on methods developed for single images and quantify model bias globally across a validation dataset, thereby facilitating inter-model comparison of bias.

Many saliency methods exist for highlighting the areas of an image most relevant to a model's prediction. Consequently, a natural validity check for saliency methods involves verifying that resulting saliency maps are compatible with known discriminatory features. Recently \cite{sanity} questioned whether saliency maps explain the predictions of a model or merely highlight the foreground of an image. In response, we propose two metrics for quantifying how accurately saliency maps reflect model behavior.

\cite{ink} identified ink skin markings as a confounder for automated diagnosis and evaluated their effect by comparing predictions on inked images to predictions following ink-removal cropping. We apply and evaluate our proposed global saliency method on a skin lesion dataset with ink artefacts.

Our key contributions are as follows:

\begin{itemize}

\item We propose a simple procedure, termed \textit{global saliency}, to aggregate saliency maps within and across images. Global saliency quantifies the effect of an artefact on a CNN's decisions across a dataset given segmentation masks for the artefact.

\item We introduce two metrics for evaluating the quality of saliency methods: model failure prediction and dataset bias detection.

\item We apply global saliency to distinguish between modes of artefact-induced model bias.

\end{itemize}

\section{Background}

\subsection{Skin Lesion Dataset}

\begin{center}
    \begin{table}[!ht]
      \caption{Training datasets and ink-co-occurrence probabilities by class}
      \label{table:cooc}
      \centering
      \begin{tabular}{llllllll}
        \toprule
        Train set     & MEL     & NV   & BCC    & SCC    & AK  & SK & Other  \\
        \midrule
        Baseline    & 58\%    &  48\%  & 74\%    & 68\%    & 76\%  & 58\% & 66\%   \\
        Unbiased   & 66\%    &  66\%  & 66\%    & 66\%    & 66\%  & 66\% & 66\%   \\
        Ink-Only    & 100\%    &  100\%  & 100\%    & 100\%    & 100\%  & 100\% & 100\%   \\
        Ablated  & 100\%    &  100\%  & 100\%    & 100\%    & 100\%  & 100\% & 100\%   \\
        \bottomrule
      \end{tabular}
    \end{table}
\end{center}

The dataset consists of 12,563 clinical skin lesion images labelled by histopathology-verified diagnoses (Appendix A). The labels are: melanoma (MEL), basal cell carcinoma (BCC), squamous cell carcinoma (SCC), actinic keratosis (AK), seborrheic keratosis (SK), nevus (NV) and ‘Other’, under which remaining less common diagnoses are grouped together. Ink markings, used to designate the location of skin biopsy, occur as a common visual artefact across the dataset (\autoref{table:cooc}, Row 1). We used a BiSeNet \cite{bisenet} semantic segmentation network to generate masks labelling the inked pixels within an image; below we denote these pixels by $A(X)$ (Appendix A). All models used in the experiments below are DenseNet-121 (Appendix A).

To address confounding via ink markings, we created three training datasets with different co-occurrence probabilities between ink and label, as shown in \autoref{table:cooc} (also, Appendix A). We compared saliency on ink across the three corresponding models to determine effects of artefact-induced bias on model predictions.

\subsection{Saliency Methods}

Given a CNN, $f$, and an input image, $X$, a \textit{saliency map} is a function, $g_f$, assigning an importance $g_f(X)_{i,j} \in \mathbb{R}$ to each pixel, $(i,j)$. A saliency map should evaluate the counterfactual importance of a subset of pixels, answering the question, "Would the model's classification change if these pixels were to be replaced by a different set of pixels?" We restrict our attention to two particular saliency maps, grad-CAM and competitive gradient$\odot$input \cite{gradcam,compet}. \textit{Grad-CAM} visualizes saliency by summing weighted filter maps within a fixed layer where the filter weight is the spatial average of its gradient. \textit{Competitive gradient$\odot$input} interprets as most salient pixels where the given class has greater absolute value than all other classes' gradients. Both grad-CAM and competitive gradient$\odot$input compute saliency with respect to a particular class, usually taken to be the model's predicted class on that image.

\subsubsection{Aggregating Saliency Across Images and the Completeness Property}

As saliency maps are constructed to compare pixels within an image, there is no guarantee that saliency values may be aggregated across images. However, certain saliency maps satisfy the \textit{completeness property}: the pixel-wise sum of saliencies equals the model's confidence for that image, $\sum g_f(X)_{i,j} = \max(f(X))$. \cite{compet} suggests that competitive gradient$\odot$input empirically satisfies the completeness property.

Given a pixel-wise segmentation of an artefact (in our case, ink), we compute saliency on the subset of pixels in which the artefact is present, $A(X) \subset X$. For saliency maps satisfying the completeness property, saliency on the artefact is normalized relative to the salience on the rest of the image (Eq. 1). Note that this normalization property does not hold for perturbative saliency methods \cite{finegrained}.

\begin{equation} \max(f(X)) - \sum\limits_{i,j \in A(X)} g_f(X)_{i,j} = \sum\limits_{i,j \in X \setminus A(X)} g_f(X)_{i,j} \end{equation}

\section{Global Quantification of Saliency}

The simplest way to aggregate saliency is to take the mean saliency, $m$, over the subset of pixels corresponding to the artefact, as defined by a semantic segmentation mask.

\begin{equation}
    m_{f,g,A}(X) = \frac{1}{\lvert A(X)\rvert}\sum\limits_{i,j \in A(X)} g_f(X)_{i,j}
\end{equation}

\begin{figure}
\centering
\begin{subfigure}{.5\textwidth}
  \centering
  \includegraphics[width=.9\textwidth]{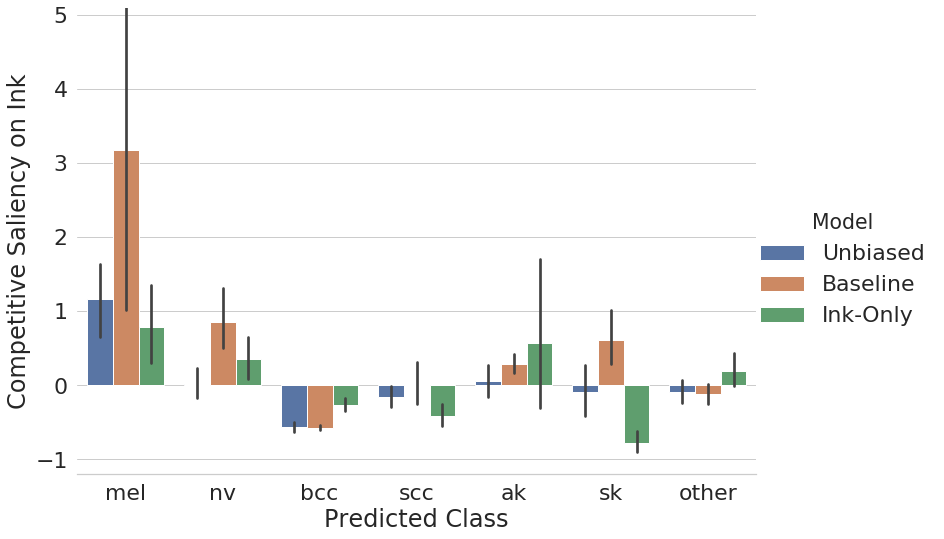}
  \caption{Global saliency using Competitive Grad$\odot$Input}
  \label{fig:global1}
\end{subfigure}%
\begin{subfigure}{.5\textwidth}
  \centering
  \includegraphics[width=.9\textwidth]{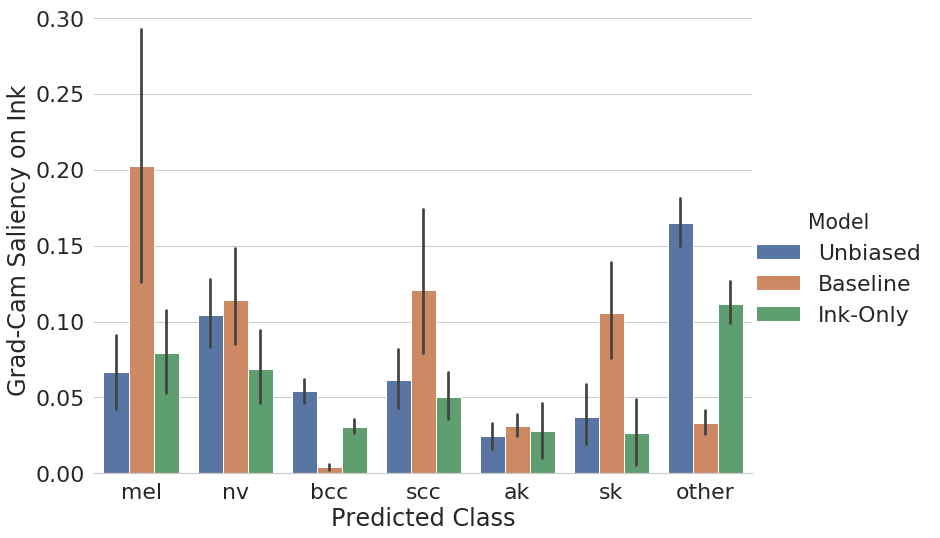}
  \caption{Global saliency using grad-CAM}
  \label{fig:global2}
\end{subfigure}
\caption{Global mean saliency on validation set, Z-score normalized by dataset. Mean and standard deviation computed over all pixels in all images, by dataset. Error bars are 95\% CI.}
\label{fig:global}
\end{figure}

The values of $m_{f,g,A}(X)$ may be aggregated across a validation set, $X_{val}$, to quantify the global effect of an artefact on the model which we term \textit{global saliency} (\autoref{fig:global}). For saliency maps which satisfy completeness, mean aggregation may be used directly. But, for methods which do not satisfy completeness, it is necessary to normalize either within the image or across the dataset. Alternatively, we define an aggregation function invariant to re-scaling of the saliency map -- e.g. rank the saliency of pixels within the image, $X$, and evaluate what percentage of the $n^{th}$ percentile of most salient pixels, $P_n(g_f(X))$, occur within the artefact, $n_{f,g,A}(X) = \frac{\lvert P_n(g_f(X)) \cap A(X) \rvert}{\lvert P_n(g_f(X)) \rvert}$.

\subsection{Evaluating Global Saliency}

\cite{sanity} showed certain saliency maps that satisfy completeness do not explain differences between trained and untrained models. For medical imaging, it is crucial that methods used to verify robustness of a model consistently detect model failure. We propose two tests for empirically evaluating the quality of saliency maps, using global saliency: (1) tracking dataset bias and (2) model failure prediction.

\textbf{Tracking Dataset Bias:} \textit{Identify the (non-)existence of undesirable bias by inferring whether the underlying training dataset has spurious correlations between an artefact and labels.} 

Given a training set with a visual artefact occurring across classes (e.g. \autoref{table:cooc}), we propose a validity check for global saliency: Is artefactual global saliency correlated with increasing dataset bias? For our skin lesion dataset, this corresponds to comparing saliency on ink across the baseline, unbiased, and ink-only datasets (\autoref{table:cooc}, Appendix A). \autoref{fig:global} shows the baseline dataset has higher inter-class ink-saliency variance relative to the unbiased and ink-only datasets -- with the exception of the `Other' class (\autoref{fig:global1}). We confirm the robustness of this conclusion by comparing \autoref{fig:global} to global saliency computed using peak saliency aggregation, $n_{f,g,A}$ instead of mean saliency, $m_{f,g,A}$ (Appendix C). The peak saliency results are visually similar for grad-CAM, but not for competitive saliency. For the baseline model, MEL and possibly NV/SK, consistently show higher saliency on ink than the other classes. Possible drivers of inter-class variance in saliency on ink are discussed in Section 4.

\begin{figure}
\centering
\begin{subfigure}{.5\textwidth}
  \centering
  \includegraphics[width=.9\textwidth]{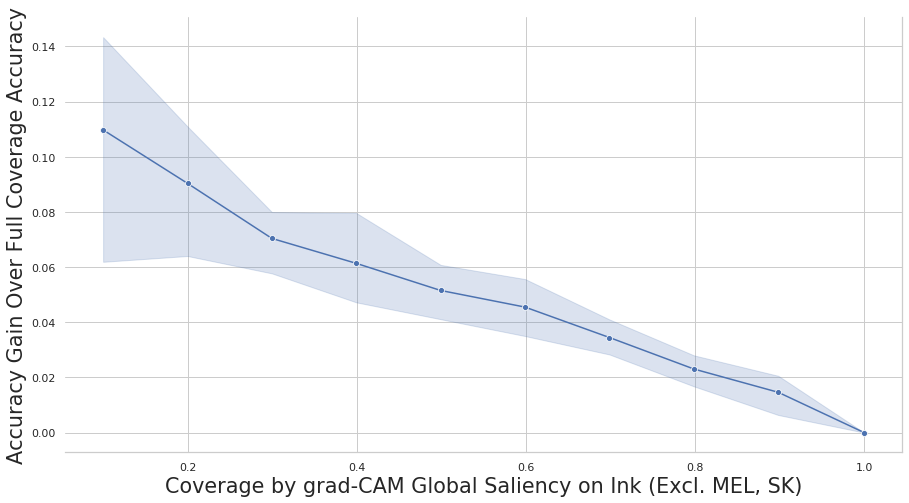}
  \caption{Val excl. MEL+SK. Kendall's $\tau$ = -0.889, \textit{P}<0.0001}
  \label{fig:corr1}
\end{subfigure}%
\begin{subfigure}{.5\textwidth}
  \centering
  \includegraphics[width=.9\textwidth]{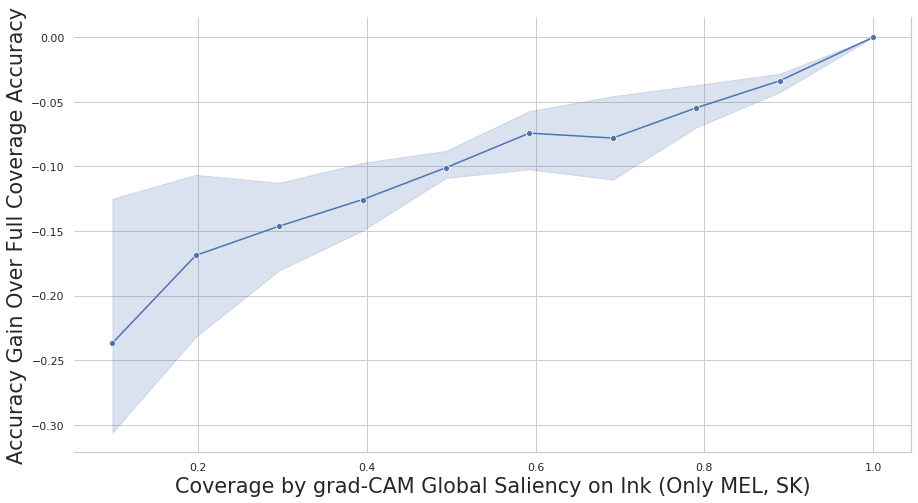}
  \caption{Val MEL+SK only. Kendall's $\tau$ = 0.806, \textit{P}<0.0001}
  \label{fig:corr2}
\end{subfigure}
\caption{3-Fold Accuracy vs ink-saliency-thresholded subsets of the validation set. Leftmost tick corresponds to the 10th percentile of images with least saliency on ink. Error margins are 95\% CI. \protect\footnotemark}
 \label{fig:corr}
\end{figure}

\textbf{Model Failure Prediction:} \textit{Predict model error due to artefact, given excessive saliency on that artefact.}

To quantify the relationship between saliency and accuracy, we evaluated the model across validation set subsets, recording model accuracy as a function of the maximum permitted saliency on ink (\autoref{fig:corr}). An ideal map maximizes the area under the response rate accuracy curve (Appendix B). 

\autoref{fig:global} and Appendix C suggest the model interprets ink as evidence for MEL and SK when trained on biased data. We test this hypothesis by calculating the correlation between model accuracy and ink saliency separately for the MEL+SK subset of the validation set and the rest (\autoref{fig:corr}). For non-MEL+SK skin lesions, excessive saliency on ink correlates with reduced accuracy whereas for MEL+SK, the baseline model spuriously interprets ink as disease-related. This relationship holds across saliency methods and aggregation schemes (Appendix B).

\footnotetext{The results shown in this figure used models trained on an expanded version of the baseline dataset for which the corresponding global saliency figures are shown in Appendix C.}

\section{Applying Global Saliency to Understand and Correct Dataset Bias}

\textbf{Understanding Dataset Bias:} Using global saliency as a measure of model bias, we compare training subsets to understand the ink bias effect on the model. Possible explanations for how ink biases the model include: \begin{enumerate*}[label={(\arabic*)}]
  \item co-occurrence rates between artefact and class label,
  \item variation in appearance between artefact instances, and
  \item visual similarity between artefact and class object.
\end{enumerate*}

It appears that co-occurrence rates drive bias in our dataset (\autoref{fig:global}). However, we have not disproven the competing hypothesis that global saliency is merely more sensitive to bias caused by co-occurrence than items 2 and 3. To quantify the effect of intra-artefact variance, we train a model solely on ink, with the lesion and skin blurred out (Appendix A). The lesion-ablated model performs no better than predicting the most prevalent class, with an accuracy of 28\%,  refuting the hypothesis that variation in ink appearance contributes to model bias. By exclusion, we propose visual similarity between skin pigment and ink as an explanation for the difference in saliency on MEL and SK compared to the other classes, but this needs to be confirmed by future experiments.

\textbf{Correcting Dataset Bias:} Since the co-occurrence unbiased model shows reduced saliency on ink, we propose training using a sampling procedure such that $P(c \mid \textit{Ink}) = P(c \mid \textit{No-Ink})$ for every class, $c$. This procedure requires no post-processing of the dataset, and \autoref{fig:global} shows that a dataset which respects this property (unbiased) reduces inter-class ink-saliency variance from $0.007$ to $0.003$; however, this variance reduction does not reach significance with Levene's test $W=1.04, P = 0.33$. It is possible that bias remains present, but is no longer detected by saliency maps. To test our sampling procedure, we evaluate the effect of lesion ablation, and these results agree with \autoref{fig:global} (Appendix A). Other solutions to model bias could involve cropping or ablating ink systematically at training time, and these augmentations could be incorporated into an unbiased sampling scheme.

\textbf{Conclusion} Global saliency allows for the quantitative evaluation of artefact-induced bias across models. The global saliency framework may also be used to evaluate the faithfulness of gradient-based saliency maps to model behavior. 

\subsubsection*{Acknowledgments}

This work was in part supported by the Helen Diller Family Comprehensive Cancer Center Impact Award and the Melanoma Research Alliance. We thank Elena Caceres for feedback on a draft of this paper. Jacob Pfau would also like to thank the Ecole Polytechnique (Palaiseau) for facilitating the internship during which he conducted this research.

\printbibliography

@MISC{dermnet,
  title        = "{DermNet} {NZ} -- All about the skin | {DermNet} {NZ}",
  booktitle    = "{DermNet} {NZ}",
  howpublished = "\url{https://www.dermnetnz.org/}",
  note         = "Accessed: 2019-9-14"
}

@ARTICLE{mednode,
  title    = "{MED-NODE}: A {Computer-Assisted} Melanoma Diagnosis System using
              {Non-Dermoscopic} Images",
  author   = "Giotis, Ioannis and Molders, Nynke and Land, Sander and Biehl,
              Michael and Petkov, Nicolai",
  journal  = "Expert Syst. Appl.",
  volume   =  42,
  number   =  19,
  month    =  may,
  year     =  2015
}

@INCOLLECTION{dermofit,
  title     = "A Color and Texture Based Hierarchical {K-NN} Approach to the
               Classification of Non-melanoma Skin Lesions",
  booktitle = "Color Medical Image Analysis",
  author    = "Ballerini, Lucia and Fisher, Robert B and Aldridge, Ben and
               Rees, Jonathan",
  volume    =  6,
  pages     = "63--86",
  month     =  jan,
  year      =  2013
}

@ARTICLE{opencv,
  title    = "The {OpenCV} library",
  author   = "Bradski, G",
  journal  = "Dr Dobb's J. Software Tools",
  volume   =  25,
  pages    = "120--125",
  year     =  2000
}

@ARTICLE{imagenet,
  title     = "Imagenet: A large-scale hierarchical image database",
  author    = "Deng, J and Dong, W and Socher, R and Li, L J and Li, K and
               {others}",
  journal   = "2009 IEEE conference",
  publisher = "researchgate.net",
  year      =  2009
}

@INPROCEEDINGS{finegrained,
  title     = "Interpretable and {Fine-Grained} Visual Explanations for
               Convolutional Neural Networks",
  booktitle = "Proceedings of the {IEEE} Conference on Computer Vision and
               Pattern Recognition",
  author    = "Wagner, Jorg and Kohler, Jan Mathias and Gindele, Tobias and
               Hetzel, Leon and Wiedemer, Jakob Thaddaus and Behnke, Sven",
  pages     = "9097--9107",
  year      =  2019
}

@ARTICLE{compet,
  title         = "A Simple Saliency Method That Passes the Sanity Checks",
  author        = "Gupta, Arushi and Arora, Sanjeev",
  month         =  may,
  year          =  2019,
  archivePrefix = "arXiv",
  primaryClass  = "cs.LG",
  eprint        = "1905.12152"
}

@INCOLLECTION{sanity,
  title     = "Sanity Checks for Saliency Maps",
  booktitle = "Advances in Neural Information Processing Systems 31",
  author    = "Adebayo, Julius and Gilmer, Justin and Muelly, Michael and
               Goodfellow, Ian and Hardt, Moritz and Kim, Been",
  editor    = "Bengio, S and Wallach, H and Larochelle, H and Grauman, K and
               Cesa-Bianchi, N and Garnett, R",
  publisher = "Curran Associates, Inc.",
  pages     = "9505--9515",
  year      =  2018
}

@ARTICLE{xray,
  title    = "Variable generalization performance of a deep learning model to
              detect pneumonia in chest radiographs: A cross-sectional study",
  author   = "Zech, John R and Badgeley, Marcus A and Liu, Manway and Costa,
              Anthony B and Titano, Joseph J and Oermann, Eric Karl",
  journal  = "PLoS Med.",
  volume   =  15,
  number   =  11,
  pages    = "e1002683",
  month    =  nov,
  year     =  2018,
  language = "en"
}

@ARTICLE{ink,
  title    = "Association Between Surgical Skin Markings in Dermoscopic Images
              and Diagnostic Performance of a Deep Learning Convolutional
              Neural Network for Melanoma Recognition",
  author   = "Winkler, Julia K and Fink, Christine and Toberer, Ferdinand and
              Enk, Alexander and Deinlein, Teresa and Hofmann-Wellenhof, Rainer
              and Thomas, Luc and Lallas, Aimilios and Blum, Andreas and Stolz,
              Wilhelm and Haenssle, Holger A",
  journal  = "JAMA Dermatol.",
  month    =  aug,
  year     =  2019,
  language = "en"
}

@ARTICLE{tomog,
  title    = "End-to-end lung cancer screening with three-dimensional deep
              learning on low-dose chest computed tomography",
  author   = "Ardila, Diego and Kiraly, Atilla P and Bharadwaj, Sujeeth and
              Choi, Bokyung and Reicher, Joshua J and Peng, Lily and Tse,
              Daniel and Etemadi, Mozziyar and Ye, Wenxing and Corrado, Greg
              and Naidich, David P and Shetty, Shravya",
  journal  = "Nat. Med.",
  volume   =  25,
  number   =  6,
  pages    = "954--961",
  month    =  jun,
  year     =  2019,
  language = "en"
}

@ARTICLE{meta,
  title    = "Accuracy of {Computer-Aided} Diagnosis of Melanoma: A
              Meta-analysis",
  author   = "Dick, Vincent and Sinz, Christoph and Mittlb{\"o}ck, Martina and
              Kittler, Harald and Tschandl, Philipp",
  journal  = "JAMA Dermatol.",
  month    =  jun,
  year     =  2019,
  language = "en"
}

@ARTICLE{densenet,
  title   = "Densely Connected Convolutional Networks",
  author  = "Huang, Gao and Liu, Zhuang and van der Maaten, Laurens and
             Weinberger, Kilian Q",
  journal = "2017 IEEE Conference on Computer Vision and Pattern Recognition
             (CVPR)",
  year    =  2017
}

@INPROCEEDINGS{bisenet,
  title     = "Bisenet: Bilateral segmentation network for real-time semantic
               segmentation",
  booktitle = "Proceedings of the European Conference on Computer Vision
               ({ECCV})",
  author    = "Yu, Changqian and Wang, Jingbo and Peng, Chao and Gao, Changxin
               and Yu, Gang and Sang, Nong",
  pages     = "325--341",
  year      =  2018
}

@INPROCEEDINGS{gradcam,
  title     = "Grad-cam: Visual explanations from deep networks via
               gradient-based localization",
  booktitle = "Proceedings of the {IEEE} International Conference on Computer
               Vision",
  author    = "Selvaraju, Ramprasaath R and Cogswell, Michael and Das, Abhishek
               and Vedantam, Ramakrishna and Parikh, Devi and Batra, Dhruv",
  pages     = "618--626",
  year      =  2017
}

@ARTICLE{health,
  title    = "Deep {Learning-A} Technology With the Potential to Transform
              Health Care",
  author   = "Hinton, Geoffrey",
  journal  = "JAMA",
  volume   =  320,
  number   =  11,
  pages    = "1101--1102",
  month    =  sep,
  year     =  2018,
  language = "en"
}

\pagebreak

\appendix

\section{Dataset and Model Details}

\subsection{Dataset Construction Details}

The skin lesion dataset consists of clinical images from \cite{dermofit,mednode,dermnet} and our institution. These datasets were aggregated into a 12,563 image \textbf{baseline} meta-dataset that was shuffled and then split 90/10 into train and validation subsets.

Typical examples of ink markings are shown below.

\begin{figure}[h!]
    \label{fig:lesions}
    \begin{minipage}[t]{0.15\textwidth}
    \includegraphics[width=\linewidth]{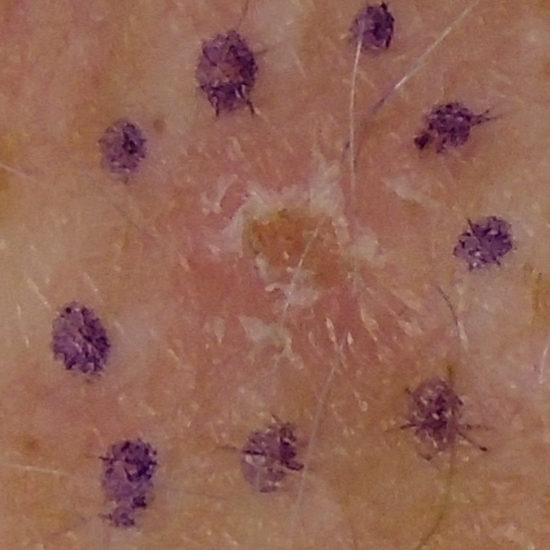}
    \end{minipage}
    \hfill
    \begin{minipage}[t]{0.15\textwidth}
    \includegraphics[width=\linewidth]{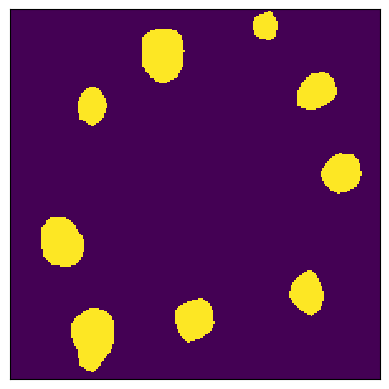}
    \end{minipage}
    \begin{minipage}[t]{0.15\textwidth}
    \includegraphics[width=\linewidth]{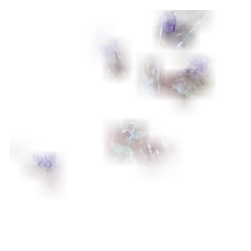}
    \end{minipage}
    \hfill
    \begin{minipage}[t]{0.15\textwidth}
    \includegraphics[width=\linewidth]{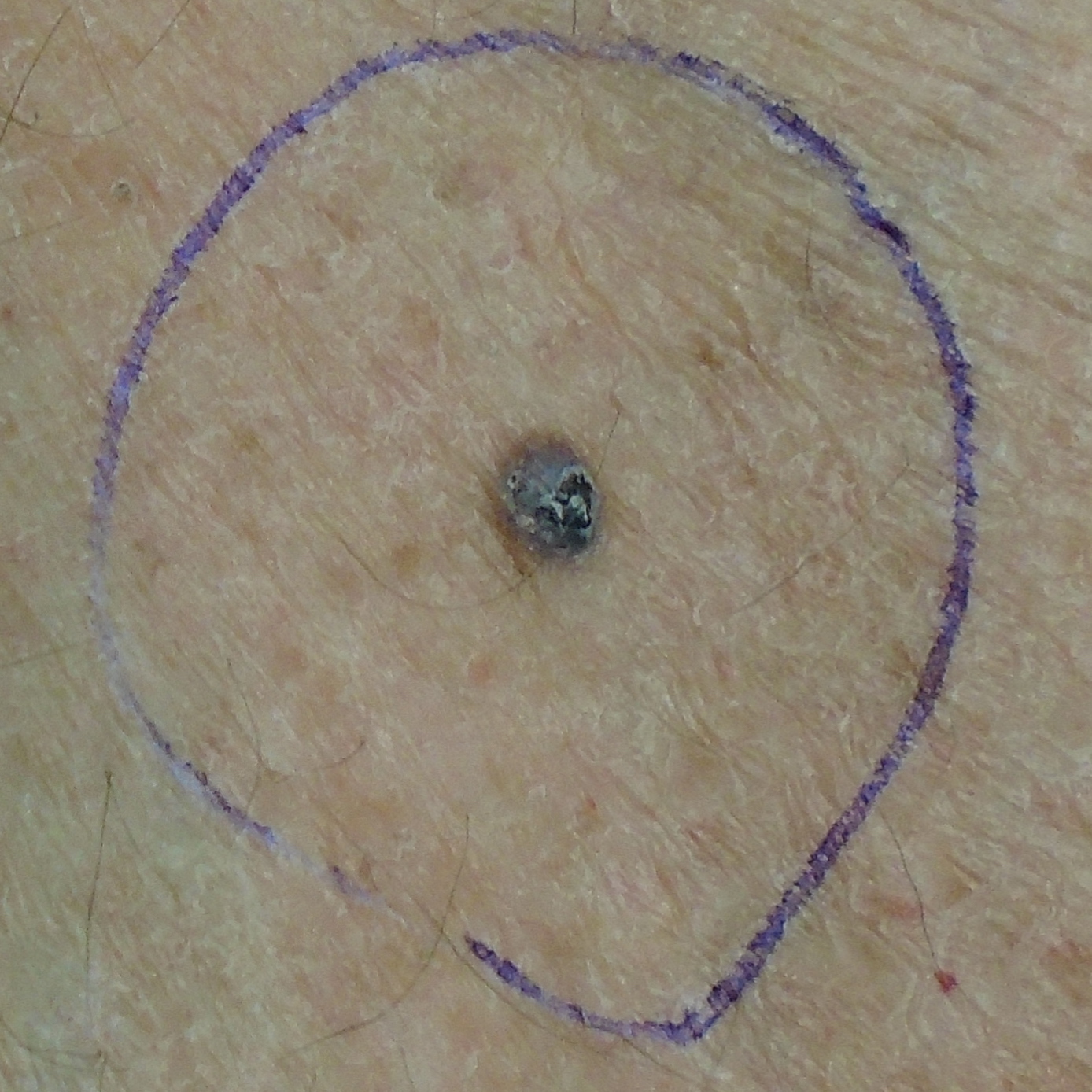}
    \end{minipage}
    \hfill
    \begin{minipage}[t]{0.15\textwidth}
    \includegraphics[width=\linewidth]{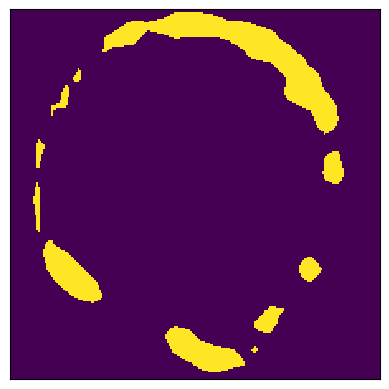}
    \end{minipage}
    \begin{minipage}[t]{0.15\textwidth}
    \includegraphics[width=\linewidth]{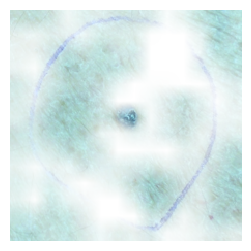}
    \end{minipage}
    \caption{Two typical samples of inked skin lesions from the dataset. From left to right: input image, BiSeNet ink segmentation mask, baseline model grad-CAM saliency overlay.}
\end{figure}

The \textbf{ink-only} dataset was constructed by using only the 8,319 images which BiSeNet predicted to contain more than 100 pixels of ink.

The \textbf{unbiased} dataset was constructed by first including the ink-only images. Then for each class, $c$, we include a randomly chosen subset $c_{uninked}$ of uninked images of that class where $\lvert c_{uninked} \rvert = 0.5*\lvert c_{inked} \rvert$. The $0.5$ ratio was chosen to maximize the dataset size while maintaining the unbiased property.

The \textbf{ablated} dataset included the same subset of images used in the ink-only dataset. These images are then ablated by replacing pixels labelled as uninked (by BiSeNet) with the mean value pixel for that image. 

As discussed in Section 4, we evaluated both the unbiased and baseline models on the ablated validation dataset. The baseline model predictions remained invariant under lesion ablation for 52\% of input images, whereas the unbiased model changed predictions to `Other' for 79\% of input images. The `Other' class may be taken as a rejection-to-predict class, and as such the behavior of the unbiased model is preferable to that of the baseline model.

\subsection{DenseNet Model Details}

All experiments and results used a DenseNet-121 architecture pre-trained on the ImageNet dataset \cite{densenet,imagenet}. The DenseNet was trained for 50 epochs using SGD with an initial learning rate of 0.02 with a 0.2 decay factor applied following 15 epochs of plateaued performance. Training used standard augmentation techniques (random flipping, rotation, color jitter, and affine transformation) as well as up-sampling of low-prevalence classes to enforce class balance.

For each model we evaluated the performance on both the inked and uninked subsets of the full validation set. In order to compare between artefact-induced bias levels between models, it is desirable to hold constant model performance. Performance on each training dataset was comparable.

\begin{center}
    \begin{table}[!ht]
      \caption{Validation set macro-mean AUROC by training set \protect\footnotemark}
      \label{table:perf}
      \centering
      \begin{tabular}{llllllll}
        \toprule
        Train set     & Full Val     & Inked Val    & Uninked Val  \\
        \midrule
        Baseline    & 0.757    &  0.744 & 0.773  \\
        Unbiased   & 0.746    &  0.746  & 0.744  \\
        Ink-Only    & 0.753    &  0.755 & 0.763  \\
        \bottomrule
      \end{tabular}
    \end{table}
\end{center}
\footnotetext{Due to time constraints, some figures in the appendices do not have confidence intervals.}

\subsection{BiSeNet Model Details}

The BiSeNet was trained using a set of 300 openCV \cite{opencv} rule-based masks selecting ink by hue and intensity. For each of these 300 images, we manually reviewed the quality of three distinct rule-based procedures and selected the mask which appeared most accurate. These 1/0 valued semantic segmentation masks were then augmented by Gaussian blurring the edges of ink regions to allow for uncertainty.

The BiSeNet was trained for 100 epochs optimized by SGD using an initial learning rate of 0.1. Learning rate was decayed by a factor of 0.2 following 12 epochs of plateaued performance. Training used standard augmentation techniques (random flipping, rotation, color jitter, and affine transformation).

Manual review of BiSeNet masks suggests that the BiSeNet performs at human level for ~95\% of images, but due to time constraints, no human-annotated validation set exists for evaluating the pixel-level performance of the BiSeNet masks.

\section{Model Failure Prediction and Response Rate Accuracy Curves}

An ideal saliency map would demonstrate when the ink within a given image is confounding the model. To quantify to what extent the saliency map succeeds in identifying the degree of confounding, we calculate the area under the response rate accuracy curve (AURRAC). The AURRAC value is equivalent to weighing the model's 1/0 loss on a given image by its percentile rank saliency on ink. A greater AURRAC indicates that the saliency method provides more accurate predictions of model failure due to ink confounding. Given a trained model, for each saliency map and global aggregation scheme we show the corresponding AURRAC. Whereas \autoref{fig:corr} showed response rate accuracy curves separated into MEL+SK subset and the rest, the following curves show the entire validation set. Due to time constraints, the following curves were computed on the original baseline dataset using one train/test split.

\begin{figure}[!ht]
\label{fig:aurraca}
\begin{subfigure}{.5\textwidth}
  \centering
  \includegraphics[width=.9\textwidth]{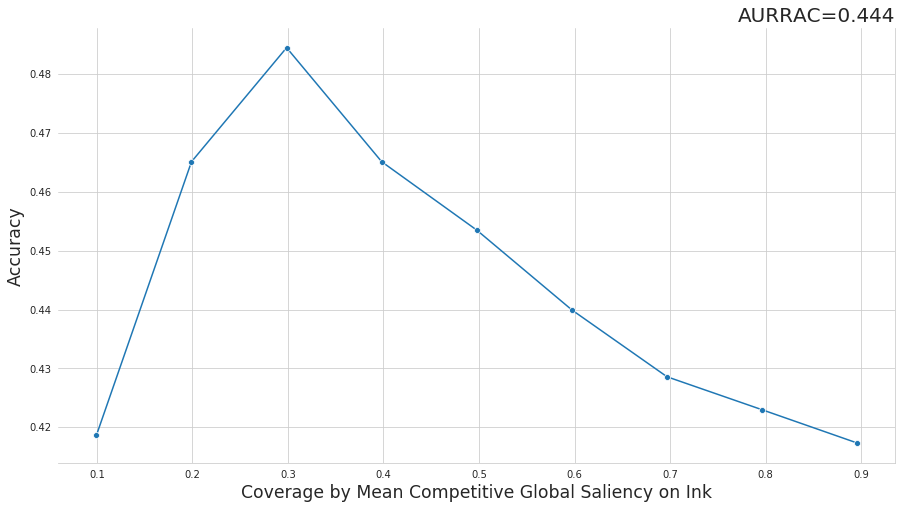}
  \caption{Mean saliency aggregation.}
  \label{fig:aurraca1}
\end{subfigure}%
\begin{subfigure}{.5\textwidth}
  \centering
  \includegraphics[width=.9\textwidth]{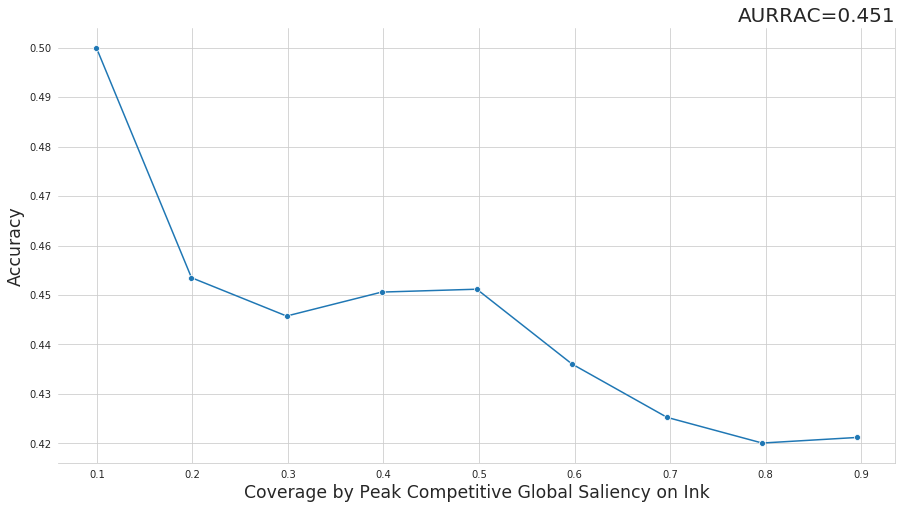}
  \caption{Peak saliency aggregation.}
  \label{fig:aurraca2}
\end{subfigure}
\caption{Accuracy vs ink-saliency-thresholded subsets of the validation set. Baseline model with Competitive Grad$\odot$Input.}
\end{figure}

\begin{figure}[!ht]
\label{fig:aurracb}
\begin{subfigure}{.5\textwidth}
  \centering
  \includegraphics[width=.9\textwidth]{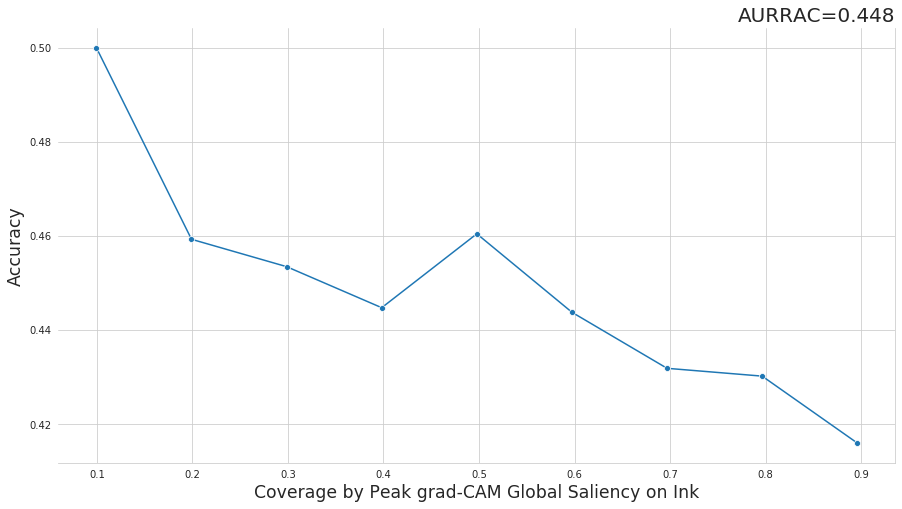}
  \caption{Mean saliency aggregation.}
  \label{fig:aurracb1}
\end{subfigure}%
\begin{subfigure}{.5\textwidth}
  \centering
  \includegraphics[width=.9\textwidth]{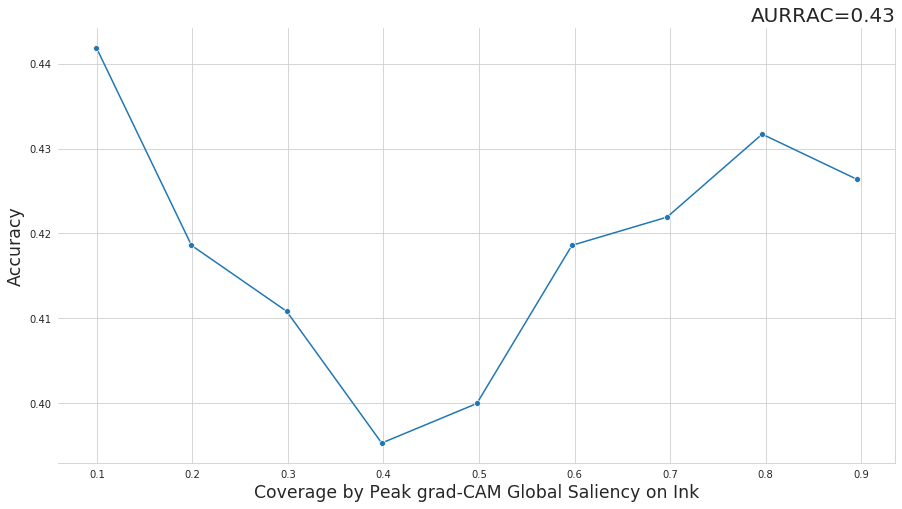}
  \caption{Peak saliency aggregation.}
  \label{fig:aurracb2}
\end{subfigure}
\caption{Accuracy vs ink-saliency-thresholded subsets of the validation set. Baseline model with grad-CAM.}
\end{figure}

These preliminary results show peak competitive saliency achieves maximal AURRAC, thereby predicting baseline model failure most consistently. In contrast, the curves shown in Figure 4a and 5b do not trend downwards indicating that these global saliency methods do not predict model failure.

\section{Dataset Bias and Global Saliency by Class}

Inter-model comparisons of global saliency using mean aggregation assume that saliency maps take on values on the same scale. For saliency maps satisfying the completeness property, any pair of models which have the same distribution of confidence values across a dataset will also have the same distribution of global saliency values. 

We show, empirically, that both grad-CAM and competitive gradient$\odot$input do not have comparable global saliency distributions across models. In particular, using competitive gradient$\odot$input, we apply the Wilcoxon signed-rank test to reject the null hypothesis that image-aggregated saliency distributions for the unbiased and ink-only models with $W=59,967, P<0.0001$. However, the Wilcoxon test cannot distinguish between the confidence distributions of the unbiased and ink-only models, $W=183,745, P=0.67$. These differences in competitive saliency distributions suggest that we must normalize saliency values for each model individually -- even when confidence levels are comparable. 

\begin{figure}[!ht]
\label{fig:conf}
  \centering
  \includegraphics[width=.6\textwidth]{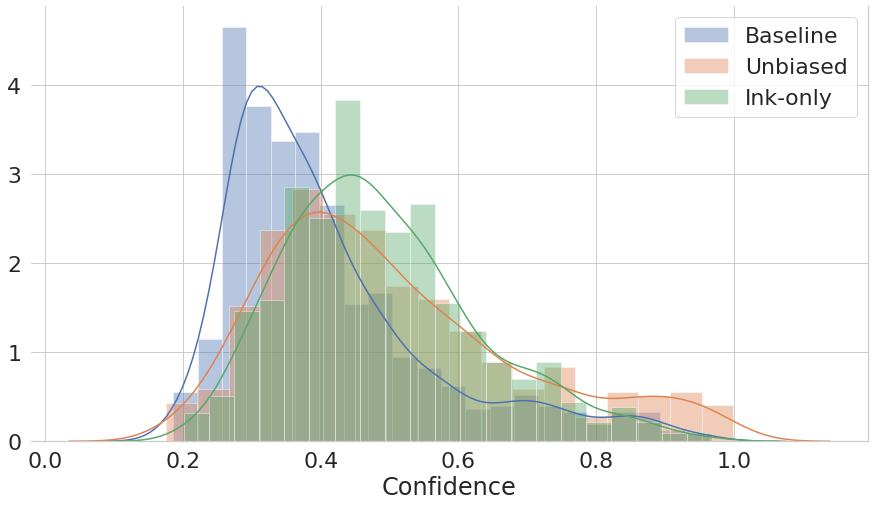}
  \caption{Validation set confidence distributions by model}
\end{figure}

\begin{figure}[!ht]
\label{fig:saldist}
\centering
\begin{subfigure}{.45\textwidth}
\label{fig:saldist1}
  \centering
  \includegraphics[width=.9\textwidth]{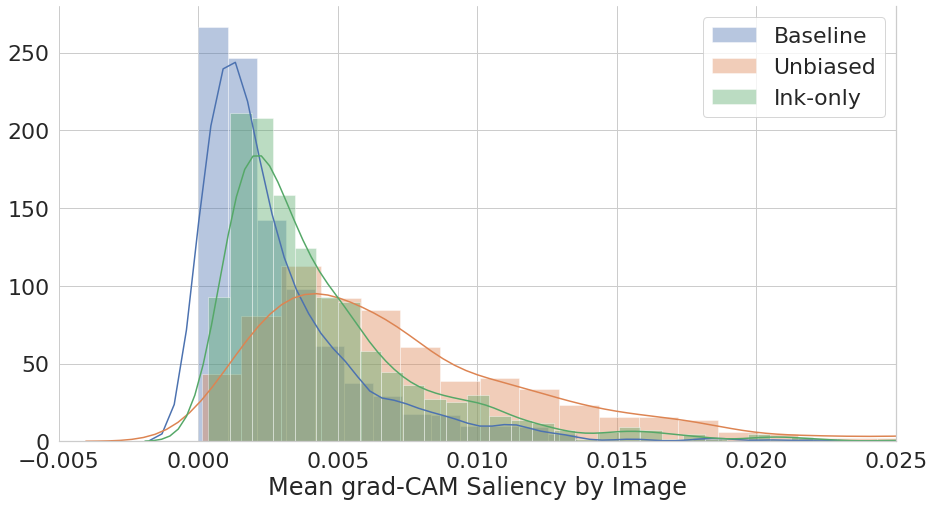}
\end{subfigure}
\begin{subfigure}{.45\textwidth}
\label{fig:saldist2}
  \centering
  \includegraphics[width=.9\textwidth]{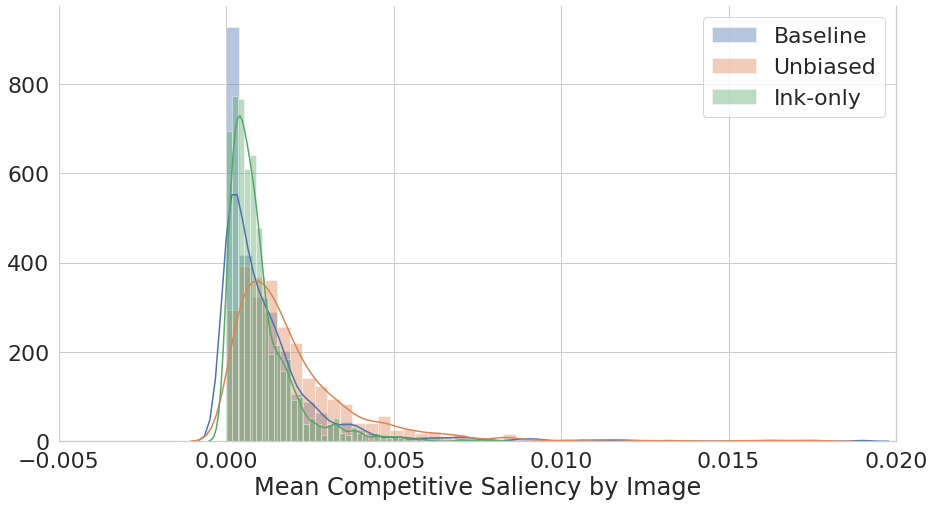}
\end{subfigure}
\caption{Validation set saliency distributions by model}
\end{figure}

We hypothesize that global saliency is largely invariant to the choice of aggregation function. Below we plot global saliency on the skin lesion dataset using the peak saliency function defined in Section 3, 

\begin{equation}
    n_{f,g,C}(X) = \frac{\lvert P_{98}(g_f(X)) \cap A(X) \rvert}{\lvert P_{98}(g_f(X)) \rvert}
\end{equation}

\begin{figure}[!ht]
\label{fig:globalb}
\centering
\begin{subfigure}{.5\textwidth}
  \centering
  \includegraphics[width=.9\textwidth]{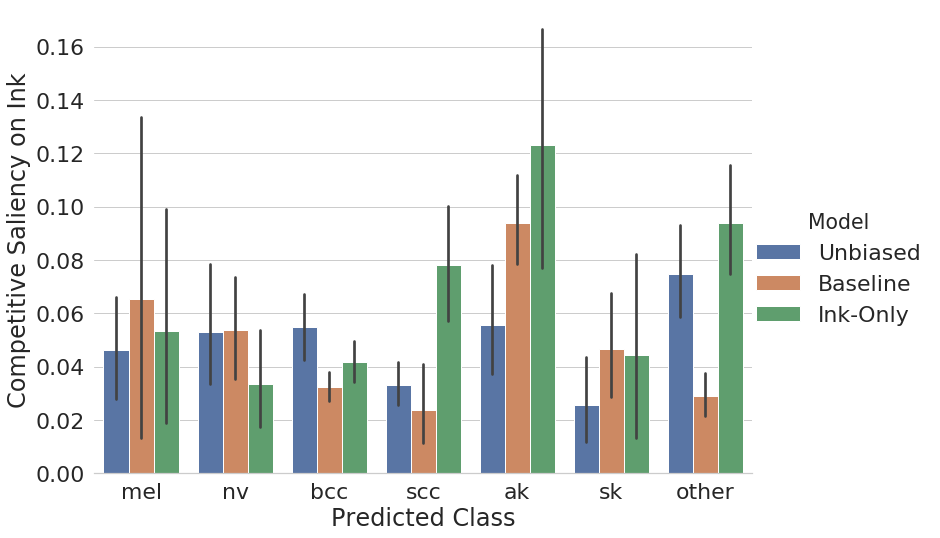}
  \caption{Global saliency using Competitive Grad$\odot$Input}
  \label{fig:globalb1}
\end{subfigure}%
\begin{subfigure}{.5\textwidth}
  \centering
  \includegraphics[width=.9\textwidth]{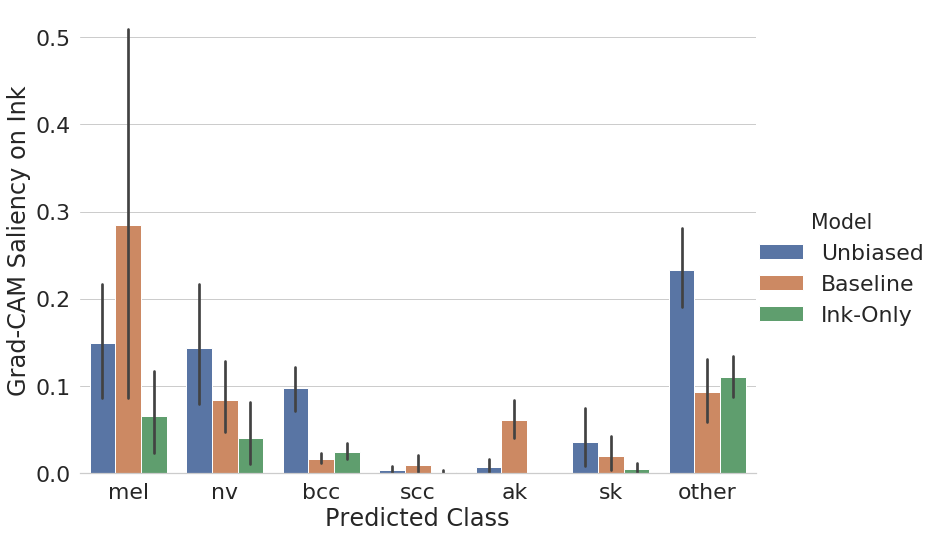}
  \caption{Global saliency using grad-CAM}
  \label{fig:globalb2}
\end{subfigure}
\caption{Global peak saliency on baseline validation set.}
\end{figure}

Note that Figure 8a does not track dataset bias, i.e. the baseline model appears to have comparable saliency on ink relative to the unbiased and ink-only models. Further research is required to determine why competitive saliency is not compatible with the $n_{f,g,C}$ function. Grad-CAM saliency appears compatible with the $n_{f,g,C}$ function, successfully tracking dataset bias. Taken together, the results of \autoref{fig:global} and  \autoref{fig:globalb2} suggest that melanoma and nevus consistently show higher saliency on ink in the baseline model (relative to the other classes).

\autoref{fig:corr} used an expanded baseline dataset (including 8,000 more recent, additional images). The global saliency by class remains similar, but the saliency on SK is higher relative to the other classes when compared to the model shown in \autoref{fig:global}. 

\begin{figure}[!ht]
\centering
  \includegraphics[width=.6\textwidth]{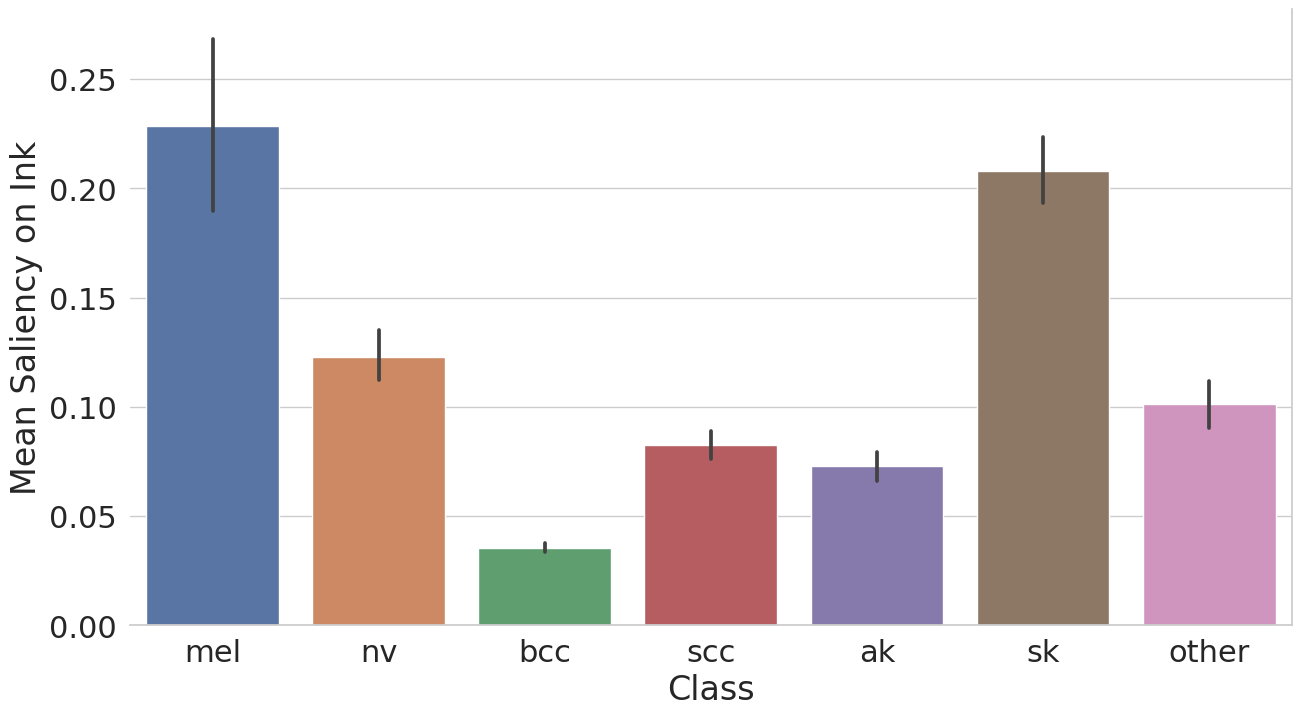}
  \caption{Expanded baseline dataset. Grad-CAM saliency aggregated across 3-folds of cross-validation.}
  \label{fig:exp}
\end{figure}%

\end{document}